\documentclass[runningheads]{llncs}
\usepackage{makeidx}
\usepackage{graphicx}
\usepackage{amsmath,amssymb}
\usepackage{color}
\usepackage[normalem]{ulem}

\begin{document}

        \pagestyle{headings}
        \mainmatter

        \def\GCPR18SubNumber{106}

        \title{DeepKey: Towards End-to-End Physical Key Replication From a Single Photograph\textcolor[rgb]{1,1,1}{.}}

        \titlerunning{DeepKey: Towards Physical Key Replication From a Single Photograph.}
        \authorrunning{Rory Smith and Tilo Burghardt}
        \author{\textcolor[rgb]{1,1,1}{-}Rory Smith \textcolor[rgb]{1,1,1}{---------------------} Tilo Burghardt\\\textsf{(rs14369@bristol.ac.uk) \textcolor[rgb]{1,1,1}{-------------} (tilo@cs.bris.ac.uk) \textcolor[rgb]{1,1,1}{----}}}
        \institute{Department of Computer Science, SCEEM, University of Bristol, UK\textcolor[rgb]{1,1,1}{--}}

        \maketitle
        
   \begin{abstract}
This paper describes DeepKey, an end-to-end deep neural architecture capable of taking a  digital RGB image of an `everyday' scene containing a pin tumbler key (e.g. lying on a table or carpet) and fully automatically inferring a printable 3D key model. We report on the key detection performance and describe how candidates can be transformed into physical  prints. We show an example opening a real-world lock. Our system is described in detail, providing a breakdown of all components including key detection, pose normalisation, bitting segmentation and 3D model inference. We provide an in-depth evaluation and conclude by reflecting on limitations, applications, potential security risks and societal impact. We contribute the DeepKey Datasets of $5,300+$ images covering a few test keys with bounding boxes, pose and {unaligned} mask~data. 
\end{abstract}
        
\section{Introduction and Overview}
\label{sec:introduction}
Imaging the detailed structural properties of physical keys is easily possible using modern high-resolution cameras or smartphones. Such photography may be undertaken by the rightful owner of a key to produce a visual backup or by an untrusted third party. The latter could potentially image keys  unnoticed,
particularly when considering scenarios that expose key rings in plain sight in public~(e.g. on a table at a cafe) or even private environments (e.g. on a kitchen counter visible through a window). 

In this paper we show that fully automated physical key generation from  photographic snapshots is a technical reality. Despite a low physical duplication accuracy experienced with the described system, this raises wider questions and highlights the need for adequate countermeasures. In particular, we will explain  here how a visual end-to-end convolutional neural network (CNN) architecture can be used to generate  3D-printable  pin tumbler key models from single RGB images without user input. To the best of our knowledge the proposed system is the first one that automates the task -- that is compared to published  semi-manual approaches of    potentially much higher replication~quality~\cite{Laxton2008ReconsideringPK}.      

\begin{figure}[t]
\centering
\includegraphics[width=350pt,height=221pt]{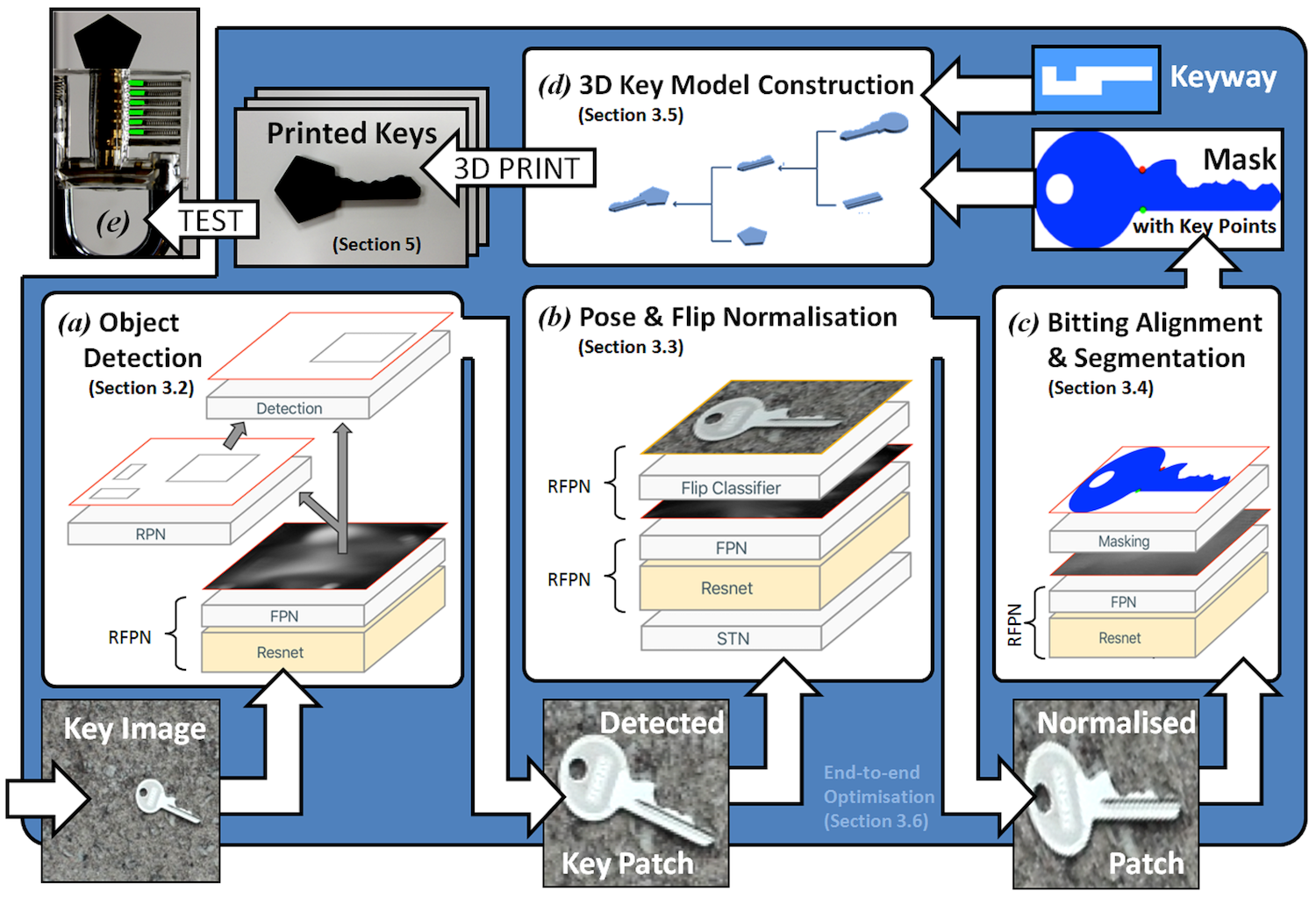}
\caption{\textbf{DeepKey Architecture.} Overview of pipeline to generate a physical pin tumbler key from an RGB photograph. \textbf{\textit{(a)}} \textit{\textbf{Key Detection: }}Images are fed to a Resnet-101-v2 \cite{He2016DeepResidual} backbone followed by a feature pyramid network (FPN) \cite{Lin2017FeaturePN}, together referred to as an RFPN component. This RFPN, a region proposal network~(RPN), and a detection network combinedly perform proposal filtering, bounding box regression and
content classification \cite{Ren2015FasterRT}. \textit{\textbf{(b) Pose Normalisation:}} Translation, scale, rotation and warp normalisation of detected key patches is then implemented via a spatial transformer network (STN) \cite{Jaderberg2015SpatialTN}, whilst a specialised RFPN controls flipping. \textit{\textbf{(c) Bitting Extraction:}} Segmentation of normalised patches is based on Mask R-CNN~\cite{He2017MaskRCNN} inferring binary bitting masks via an extended RFPN. \textit{\textbf{(d) Key  Inference:}} Masks are finally converted to 3D-printable models to yield plastic prints of  Yale keys. \textbf{\textit{(e)~Physical Tests:}} Given models of sufficient quality, these key prints may open real-world locks, although many {distinct models} may be required. }
\label{fig:overview}
\end{figure}

As depicted in Figure \ref{fig:overview}, the deep neural network (DNN) pipeline put forward here takes an `everyday' scene containing a Yale pin tumbler key as input. It then breaks vision-based key reconstruction down into a sequence of distinct inference tasks: first, keys are detected in the scene via a pose-invariantly trained Faster R-CNN \cite{Ren2015FasterRT} component, whose outputs are transformed into a unified pose domain by a customised spatial transformer network (STN) \cite{Jaderberg2015SpatialTN}. These pose-normalised patches containing aligned instances are used to infer the bitting pattern, exploiting the concepts of Mask R-CNN \cite{He2017MaskRCNN}. 

Finally, alignment and projection of the bitting mask onto a known `keyway' yields a CAD entity, which may be 3D-printed into a physical object. Using portions of the DeepKey Datasets as training information, our network pipeline is first optimised at the component level before final steps proceed in a forward-feeding end-to-end manner allowing for optimisation by the discovery of  weight correlations.

Before describing our architecture,  training and recorded performance in detail, we briefly review methodologies and prior work relevant to the application.

\section{Related Work and Context}
\label{sec:background}
    
\subsection{Vision-based Key Replication }    
\begin{figure}[t]       
\centering
\includegraphics[width=82pt,height=50pt]{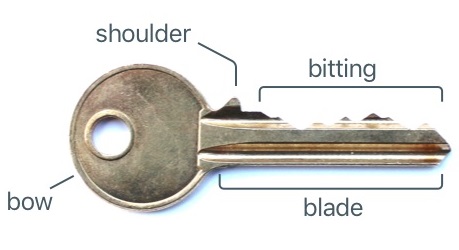} 
\includegraphics[width=262pt,height=50pt]{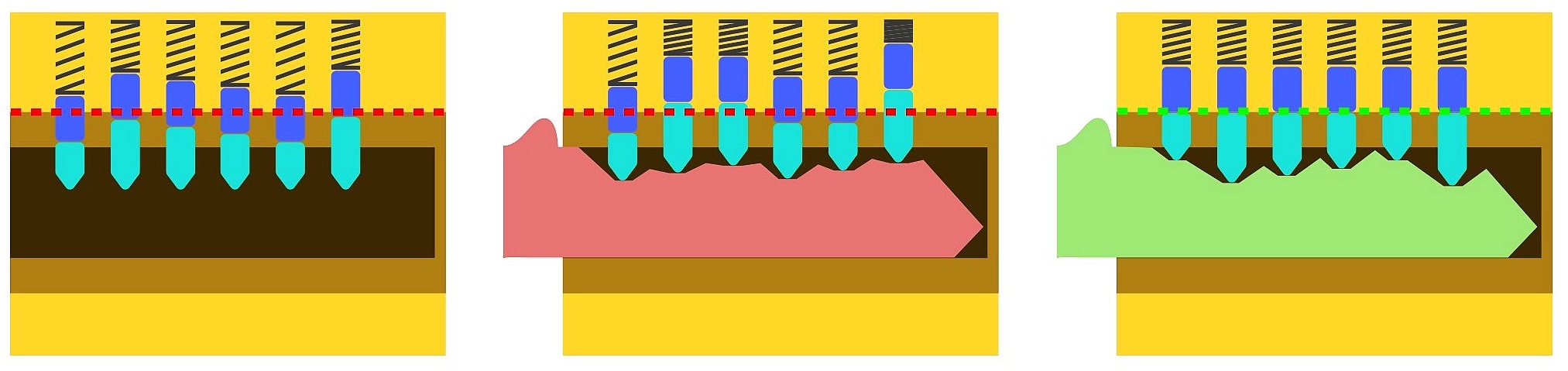}\\
\textbf{\textit{(a)}}\textcolor[rgb]{1,1,1}{..........................}\textbf{\textit{(b)}}\textcolor[rgb]{1,1,1}{...............................}\textit{\textbf{(c)}}\textcolor[rgb]{1,1,1}{...............................}\textit{\textbf{(d)}}
\caption{\textbf{Pin Tumbler Keys and Locks.} \textbf{\textit{(a)}} \textit{\textbf{Pin Tumbler Keys: }}For a given manufacturer type the non-public key information is encoded in the bitting, which is the target of the visual reconstruction. A key's keyway is a secondary level of security, but is publicly available, given the key type is known. Allowed cuts adhere to a maximum adjacent cut specification (MACS) to ensure insertion and extraction of the key is always possible. \textbf{\textit{(b-d)}} \textit{\textbf{Pin Tumbler Locks: }}Unlocking requires raising a set of stacked pins to particular, key-specific heights such that the entire cylinder can rotate cleanly.}
\label{fig:keys}
\end{figure}

For this proof of concept we focus on the vision-based replication of Yale pin tumbler keys only, a widely used key class and lock arrangement. As shown in Figure~\ref{fig:keys}, pin tumbler locks require the key to raise a set of stacked pins at different heights such that the entire plug may rotate cleanly. This is achieved by cuts made into one edge of the key, known as the `bitting'. For a given pin tumbler lock type, such as Yale, it is the bitting alone that encodes the information unique to an individual key. A key type's remaining geometric information including its `keyway' frontal profile is publicly available via manufacturer type patents that legally prevent the reproduction of uncut keys without a license. Even without accurate schematics, a key's frontal profile can be determined from a single photo of the key's lock visible from the outside of the door \cite{Burgess2015ReplicationPA}.
 
\textbf{Existing Vision-based Key Replication.} Computational teleduplication of physical keys via optical decoding was first published by Laxton et al. \cite{Laxton2008ReconsideringPK} who designed a semi-automated system named `Sneakey'. Their  software requires the user to crop the key from an image and annotate manually two separate  point sets: a  key-type-dependent one enabling planar homographic normalisation to a rectified pose, and a lock-dependent one for decoding the individual bitting~code. 
\subsection{Deep Learning Concepts}
In this paper, we assume the key type {to be} fixed  and the keyway to be known. Automating vision-based key replication based on this, then evolves around solving three  `classic' vision tasks:  that is object(-class) detection and localisation~\cite{AlexKrizhevsky2012ImageNetCW} of keys, rigid key pose estimation and normalisation \cite{Jaderberg2015SpatialTN}, as well as image segmentation \cite{He2017MaskRCNN} to extract the bitting pattern. A projection of the bitting onto the known keyway will then generate a scale-accurate description as a CAD model. Each of these three tasks has its own long-standing history, whose review is beyond the scope of this paper. Thus, we focus on the most~relevant~works~only.\\
\label{sec:literature_review:machine_learning:object_detection}
\textbf{Object Detectors.} For the pipeline at hand, mapping from input images to localisations of instances is fundamental to spatially focus computational attention. Across benchmark datasets \cite{pascal-voc-2012,AlexKrizhevsky2012ImageNetCW,coco} neural architectures now consistently outperform traditional vision techniques in both object detection as well as image classification \cite{deformable,AlexKrizhevsky2012ImageNetCW,yolo9000,Simonyan2014VeryDeepC}. Region-based convolutional neural networks (R-CNNs) \cite{Girshick2014RCNN} combine these  tasks by unifying candidate localisation and classification -- however, in its original form, R-CNNs are computationally expensive~\cite{Girshick2015FastRCNN}. By sharing
operations across proposals as in Fast R-CNN or SPPnet \cite{Girshick2015FastRCNN,HeKaimingSpatial} efficiency can be gained, although proposal estimation persists to be a
bottleneck. To address this, Ren et al. introduced region proposal networks (RPNs) \cite{Ren2015FasterRT}, which again share features during detection, resulting in the Faster R-CNN \cite{Ren2015FasterRT}. We use this approach here for initial candidate key detection, noting that various alternative architectures such as YOLO \cite{yolo9000} and Overfeat \cite{Sermanet2013OverFeatIR} are also viable.\\
\textbf{Network Substructure.} The base component in Faster R-CNNs, also referred to as its `backbone', can be altered or exchanged without breaking the approach's conceptual layout. Practically,  deeper backbone networks often lead to improved detection performance \cite{Simonyan2014VeryDeepC}. In response to this observation and as suggested by He et al. \cite{He2017MaskRCNN}, we utilise the well-tested 101-layer Resnet-101-v2 \cite{He2016DeepResidual} backbone in our work with final weight sharing (see Section \ref{sec:implementation_and_methodology:end_to_end_implementation}). As shown in Figure \ref{fig:archdet}, its output feeds into a feature pyramid network (FPN) \cite{Lin2017FeaturePN} similar in spatial layout to traditional scale-space feature maps \cite{Adelson1984PyramidMI}. When considering the detection application at hand where key height and width can vary vastly, the explicit use of an FPN helps to detect keys at these various scales \textit{fast}. Moreover, together with the backbone, it forms a versatile  RFPN network pair. \\
\textbf{Normalisation and Segmentation.} Fundamentally, RFPN components learn scale-space features and, thus, have been shown to support a versatile array of mapping tasks \cite{Ren2015FasterRT,He2017MaskRCNN} -- including segmentation. Mask R-CNNs \cite{He2017MaskRCNN} exemplify this practically by adding further convolutional and then de-convolutional layers to an RFPN in order to map to a binary mask (see Figure \ref{fig:archseg}). We base our segmentation architecture off this concept, but apply  modifications to increase the final mask resolution (see Section \ref{sec:implementation_and_methodology:semantic_segmentation}). Additionally, spatial transforms arising from variance in viewpoint are rectified. Spatial transformer networks (STNs)~\cite{Jaderberg2015SpatialTN} are designed to deal with this task, although traditionally in an implicit way, where the parameters of the transformation matrix that affects the image are not known. Our STN estimates transformation matrix parameters directly after object detection, unifying the representation and production of orthonormal key views for segmentation (see Figure \ref{fig:archnorm}).   
 
\section{DeepKey Implementation}
\label{sec:main_body}

\subsection{Generation of Training Information}
\label{sec:data_collection:datasets}
\begin{figure}[t]
\centering
\includegraphics[width=172pt,height=130pt]{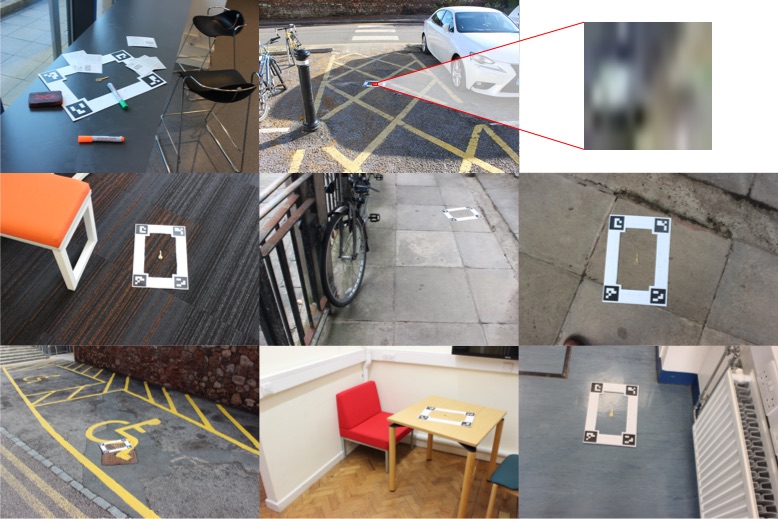} \includegraphics[width=172pt,height=130pt]{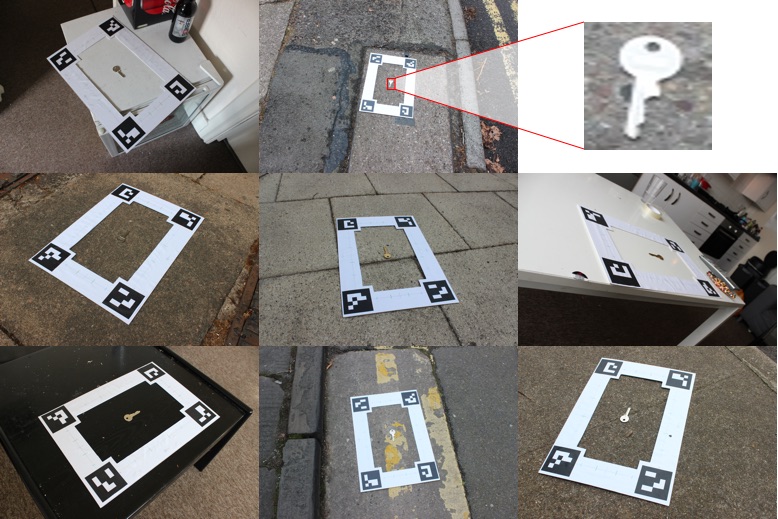}
\caption{\textbf{DeepKey Datasets.} \textbf{\textit{(left)}} \textit{\textbf{Lower Resolution Dataset A: }}Examples from $2,653$ `everyday' scenes resolved at 5184x3456 pixels containing both rectification patterns for training and an ordinary pin tumbler key at resolutions ranging from $78\times21$ to $746\times439$ primarily used to train key detection components;  \textbf{\textit{(right)}} \textit{\textbf{Higher Resolution Dataset B: }}Examples from $2,696$ scenes of seven different pin tumbler keys and rectification patterns resolved at 5184x3456 pixels with key resolutions from $171\times211$ to $689\times487$ pixels primarily used to train bitting segmentation networks. {Lowest resolution key patches from both datasets are visualised in the upper-right-most images.} }
\label{fig:datasets}
\end{figure}

\begin{figure}[t]
\centering
\includegraphics[width=172pt,height=130pt]{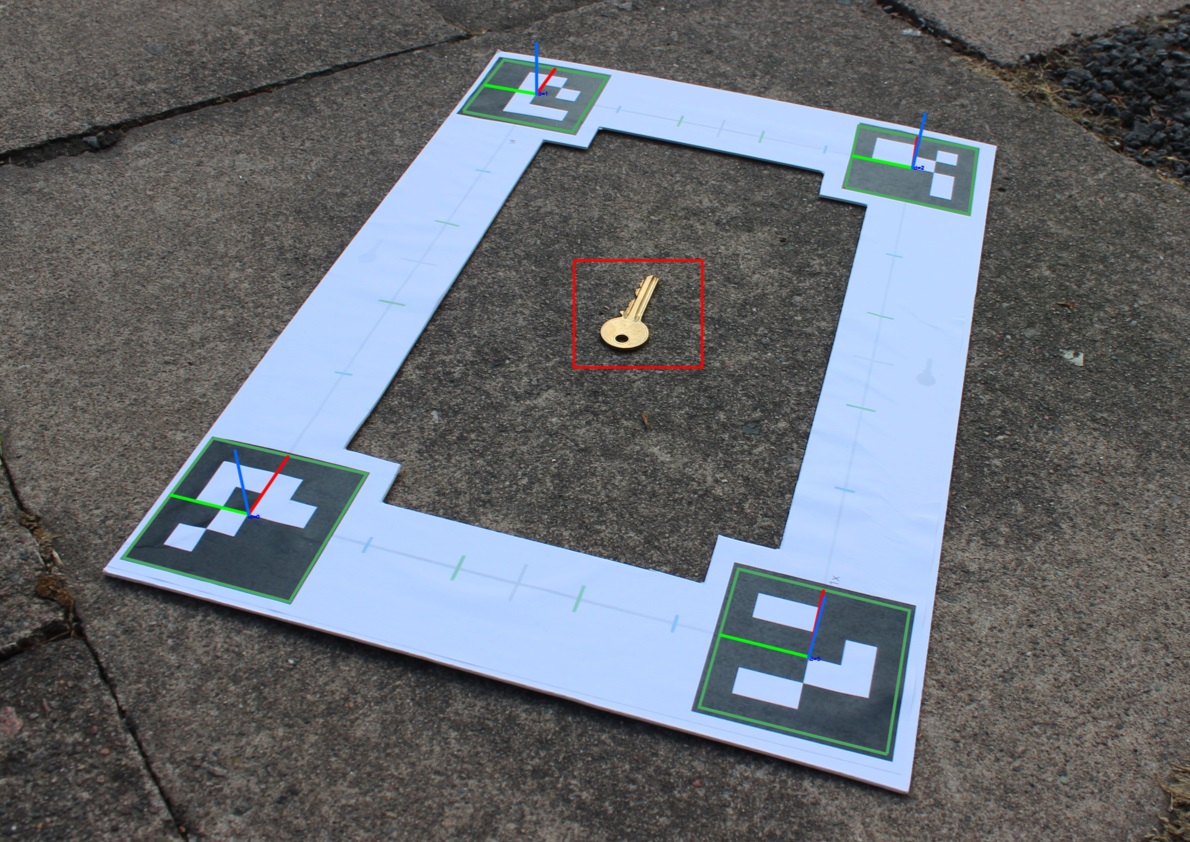} \includegraphics[width=172pt,height=130pt]{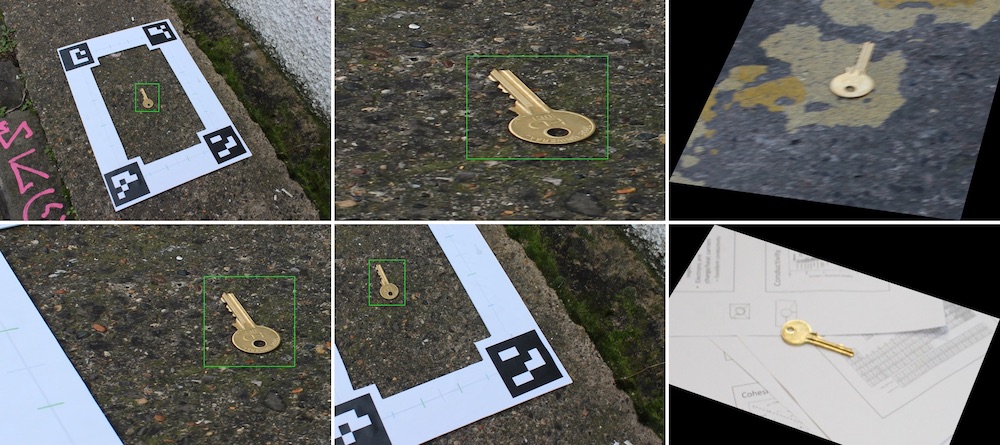}
\caption{\textbf{ Perspective Ground Truth and Augmentation.} \textbf{\textit{(left)}} \textit{\textbf{Marker Frame: }}Annotation of training images via 4 ArUco markers, pasted on a rigid card base. Automatic labelling locates each marker and generates a bounding box and transformation matrix per image. {\textbf{\textit{(right)}} \textit{\textbf{Synthetic Data Augmentation: }}The application of pseudo-random crop, scale, flip and shift augmentations to the data yields further synthetic data. Shown are an original example image (top left) and three derived augmentations next to it.  In addition, the two right-most images illustrate the marker frame exclusion augmentation used as an alternative evaluation strategy (see Section~\ref{sec:experiments}).} }
\label{fig:marker}
\end{figure}

We collected $5,349$ images of a few keys  provided in two separate DeepKey Datasets\footnote{ DeepKey Datasets can be requested  via  \texttt{https://data.bris.ac.uk}}~A and B, detailed in Figure \ref{fig:datasets}, plus a tiny extra Test C (see Section~\ref{sec:poc}) featuring a key not contained in either set. Set A contains distant shots of common environments that may contain keys -- such as tables, carpets, road surfaces, and wooden boards. 
Set B contains shots with keys  at higher resolution, but less environmental context. We divide the datasets into a traditional split of approx. 70\% for system training and a withheld portion of remaining images for validation. To allow for automatic data annotation, we developed a physical marker frame which is placed around keys in the real world (see Figures \ref{fig:datasets} and~\ref{fig:marker}), with placement aided by a custom-developed mobile app using an iPhone 7 smartphone. The marker board essentially provides four points, which can be used during a postprocessing step to calculate a projective transform per image. This automatic meta-annotation allows us to reduce labelling and produce larger training sets with exact parameterisation for backpropagation  during learning.       

    \subsection{Pose-invariant Key Detection}
    \label{sec:implementation_and_methodology:object_detection}
\begin{figure}[t]
\centering
\includegraphics[width=172pt,height=130pt]{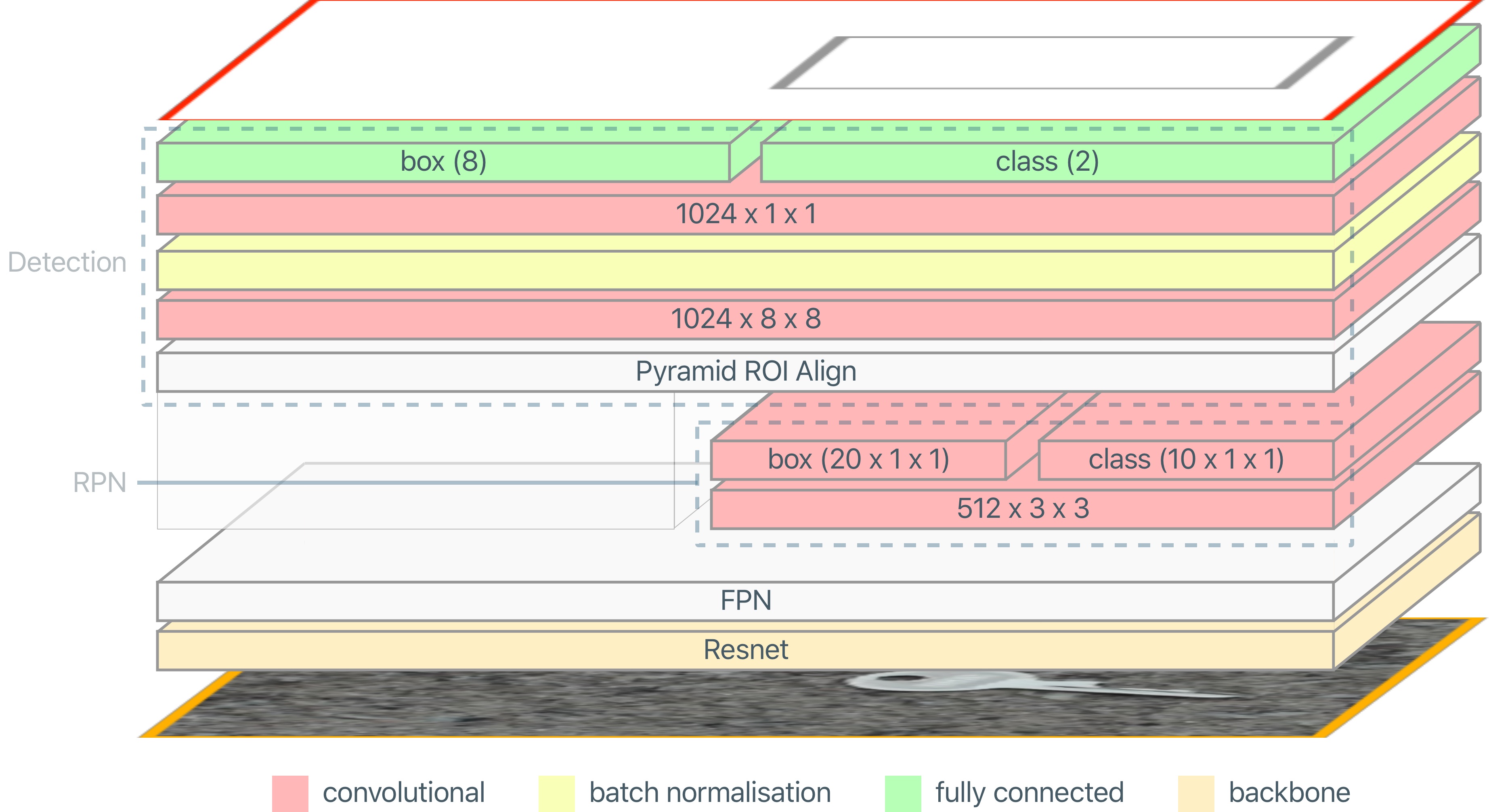} \includegraphics[width=172pt,height=130pt]{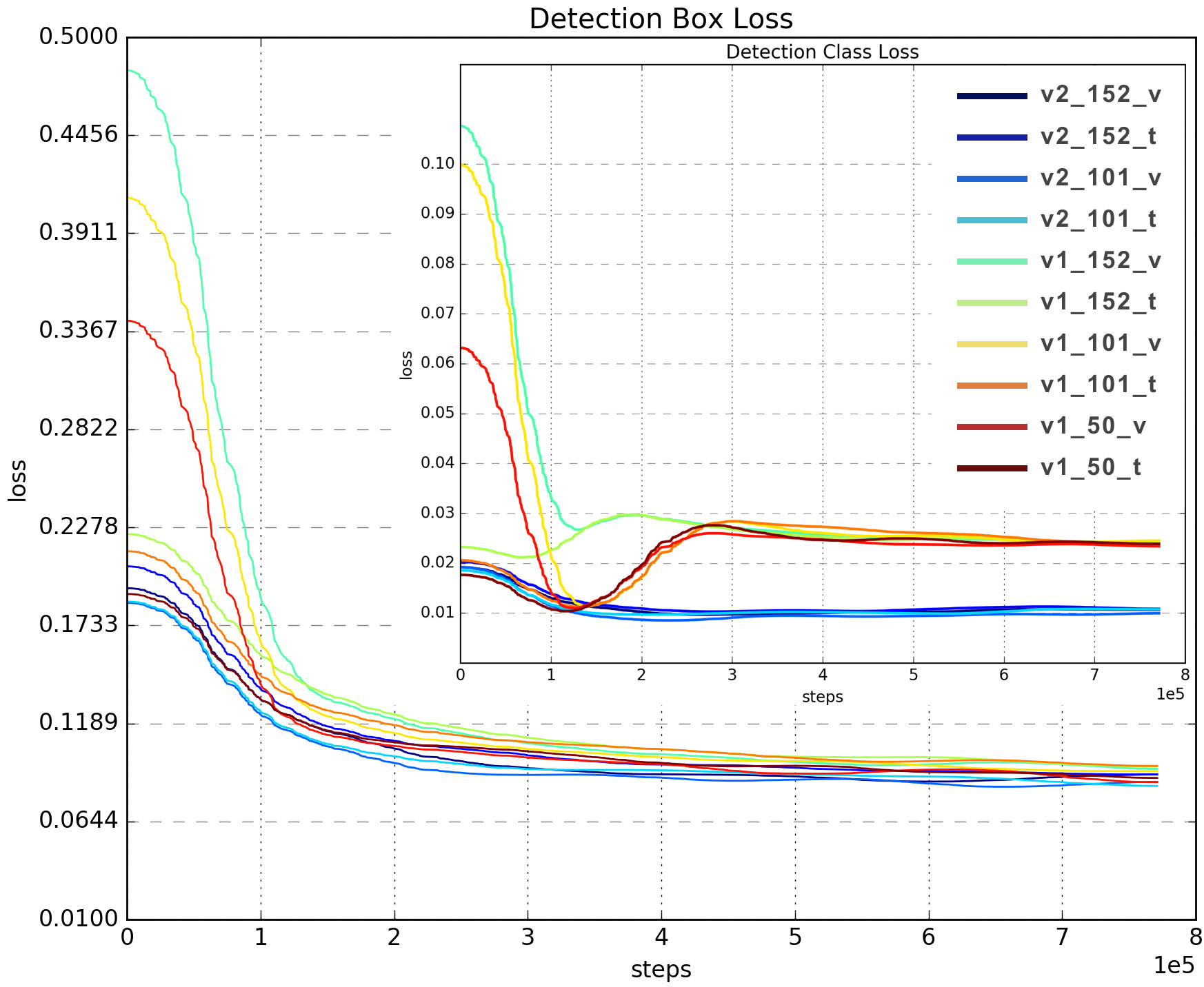}
\caption{\textbf{Detection Network Architecture.} \textbf{\textit{}}  \textbf{\textit{(left)}} \textit{\textbf{Component Details: }}In-depth description of the layout of our detection component, where Resnet-101-v2 is used as the backbone. \textbf{\textit{(right)}} \textit{\textbf{Backbone Performance Study: }}Results depict the performance impact of using different networks as backbone. Note the performance/size trade-of struck by Resnet-101-v2 compared to the various alternatives investigated.}
\label{fig:archdet}
\end{figure}

 Our detailed detection architecture is depicted in Figure \ref{fig:archdet}. Woven around a Faster R-CNN layout with a training RPN proposal limit of $200$ we utilise a Resnet-101-v2 as the network backbone. An empirical study of similar alternatives as shown in Figure \ref{fig:archdet} (right) confirms its efficacy. Inspired by Mask R-CNN~\cite{He2017MaskRCNN}, the FPN fed by this backbone provides improved scale-dependent detection performance. It takes Resnet blocks $C2, C3, C4, C5$ and adds lateral connections as shown in Figure \ref{fig:archdet}. We feed in resized images at $224\times224$ pixels and used a pooling size of $8\times8$ when resizing cropped feature-map regions. With training images augmented from DeepKey Datasets A and B, we first froze our backbone to fine-tune remaining layers via SGD with momentum ($\lambda = 0.9$) with batch size 16 and learning rate (LR) of $0.001$ for $400$ epochs -- before  unfreezing the backbone and lowering the LR by factor $10$ and optimizing for  $400$ epochs further. We use log losses for RPN and detection head classification as detailed in \cite{Ren2015FasterRCNN}, and a smooth L1 loss as defined in \cite{Girshick2015FastRCNN} for bounding box regression. We evaluate the detection component in detail in Section \ref{sec:experiments} where Figure \ref{fig:resdet} visualises results. Note that we carry out a comparative study confirming that the potential presence of marker frame pixels in the receptive fields of network layers has no application-preventing impact on key detection performance.

\begin{figure}[!hb]
\centering
\includegraphics[width=116pt,height=130pt]{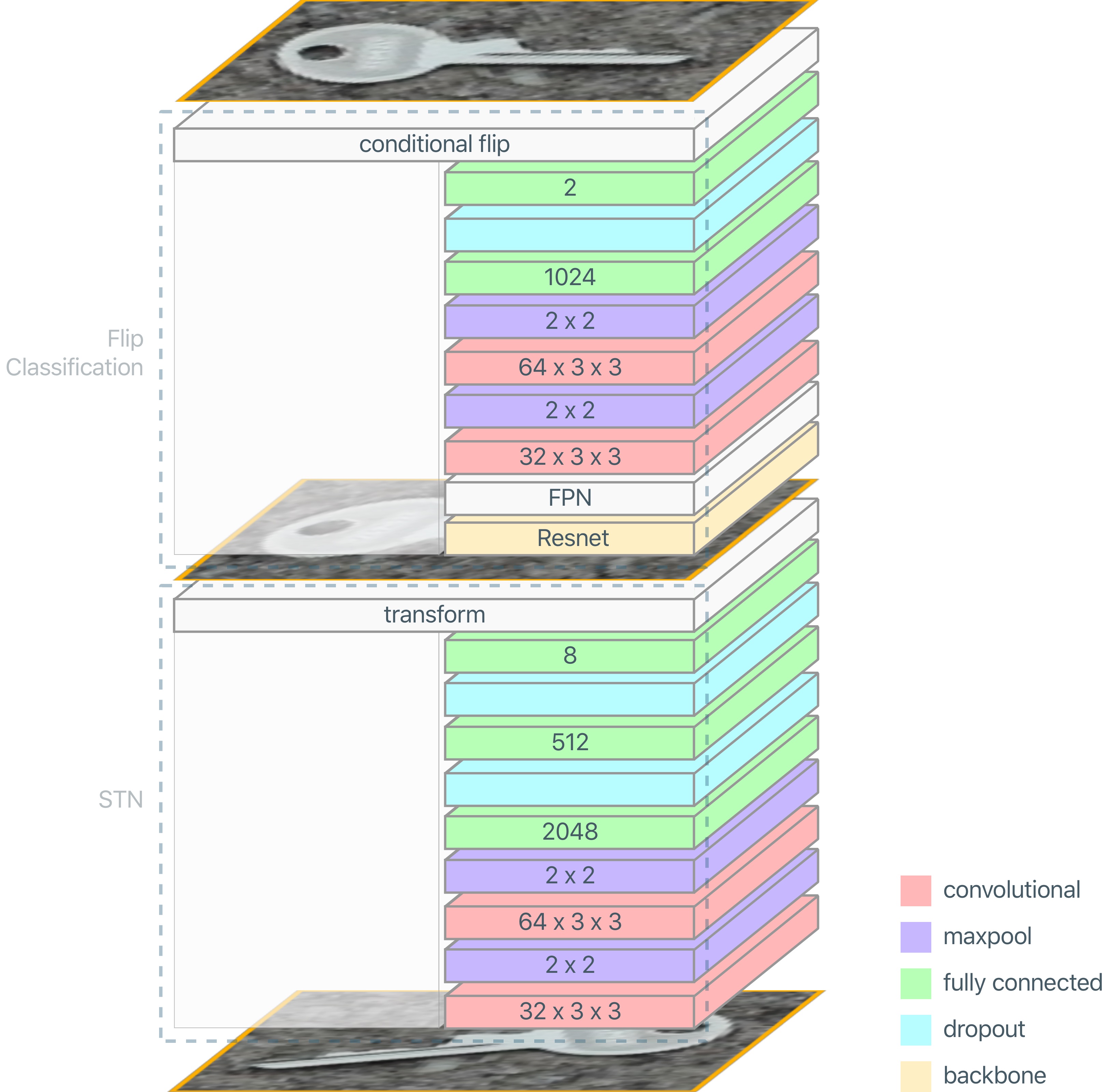} \includegraphics[width=228pt,height=130pt]{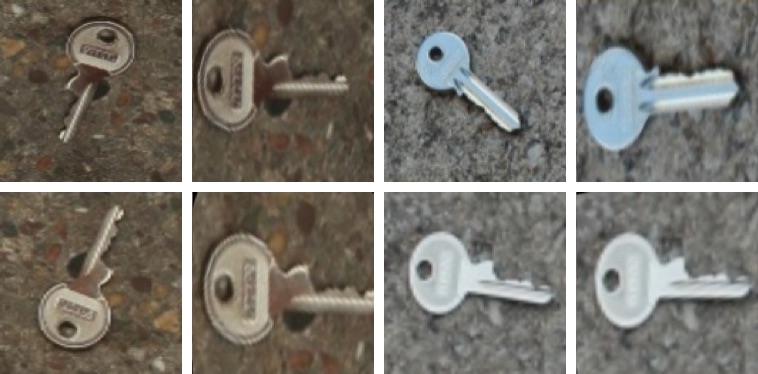}
\caption{\textbf{Pose Normalisation Architecture.} \textbf{\textit{(left)}} \textit{\textbf{Component Details: }}In order to normalise pose, an STN predicts 8 mapping parameters on whose basis a perspective transform is applied (white). Subsequently, a classification network determines whether or not an image flip should be performed. \textbf{\textit{(right)}} \textit{\textbf{Examples of Normalised Keys: }}Representative sample patches before and after pose normalisation and flip correction. }
\label{fig:archnorm}
\end{figure}

     \subsection{Pose Normalisation}
    \label{sec:implementation_and_methodology:image_normalisation}
     In order to map key detections into unified pose we opt to use traditional geometric transform operations fuelled by deep network predictions of eight perspective transform parameters as well as flips. Figure \ref{fig:archnorm} (left) shows the used architecture in detail. Following {\cite{zakka_2017}}, an STN first predicts eight parameters forming a transformation matrix, which are applied to the input via a projective transform. Rather than applying STNs in a traditional unsupervised manner, we train against our labelled dataset of transform parameters, using an averaged L2 norm loss (MSE) detailed in \cite{SchmidhuberJuergenDeepLearning}. Note that we do not use the features from our Resnet-101 backbone, {but} instead opt to take a shallower number of convolutions from the raw image.
    A subsequent flip classification network determines whether the key requires flipping such that its bitting faces up, readily aligned for bitting segmentation. Using SGD with momentum ($\lambda=0.9$) and batch size 32, pose normalisation and flip classification are repeatedly trained on DeepKey Dataset B  key patches, with random augmentations resolved at  $128\times128$ pixels, against pre-computed marker frame ground truths as detailed in Figure~\ref{fig:marker}. \\ Inspired by~\cite{Jaderberg2015SpatialTN}, pose normalisation uses a LR of $0.01$ and flip classification a LR of $0.00001$ for the first $400$ epochs. Afterwards, LRs are reduced by  $10\times$ before continuing for further $400$ epochs. {Flip classification uses a softmax cross entropy loss as in \cite{SchmidhuberJuergenDeepLearning}.}
    As in the detection network, all non-final convolutional and fully connected layers use ReLU as non-linearity. Only fully-connected layers are regularised via dropout at an empirically optimised rate of $0.3$. 
    During pose normalisation training parameters of the loss are normalised. Note that the standard deviation of each parameter is found and based on that the eight free parameters $\theta_{0}, \theta_{1}, \theta{_2}...\theta_{7}$ are normalised to speed up convergence.

\subsection{Bitting Segmentation}
\label{sec:implementation_and_methodology:semantic_segmentation}
\begin{figure}[t]
\centering
\includegraphics[width=117pt,height=95pt]{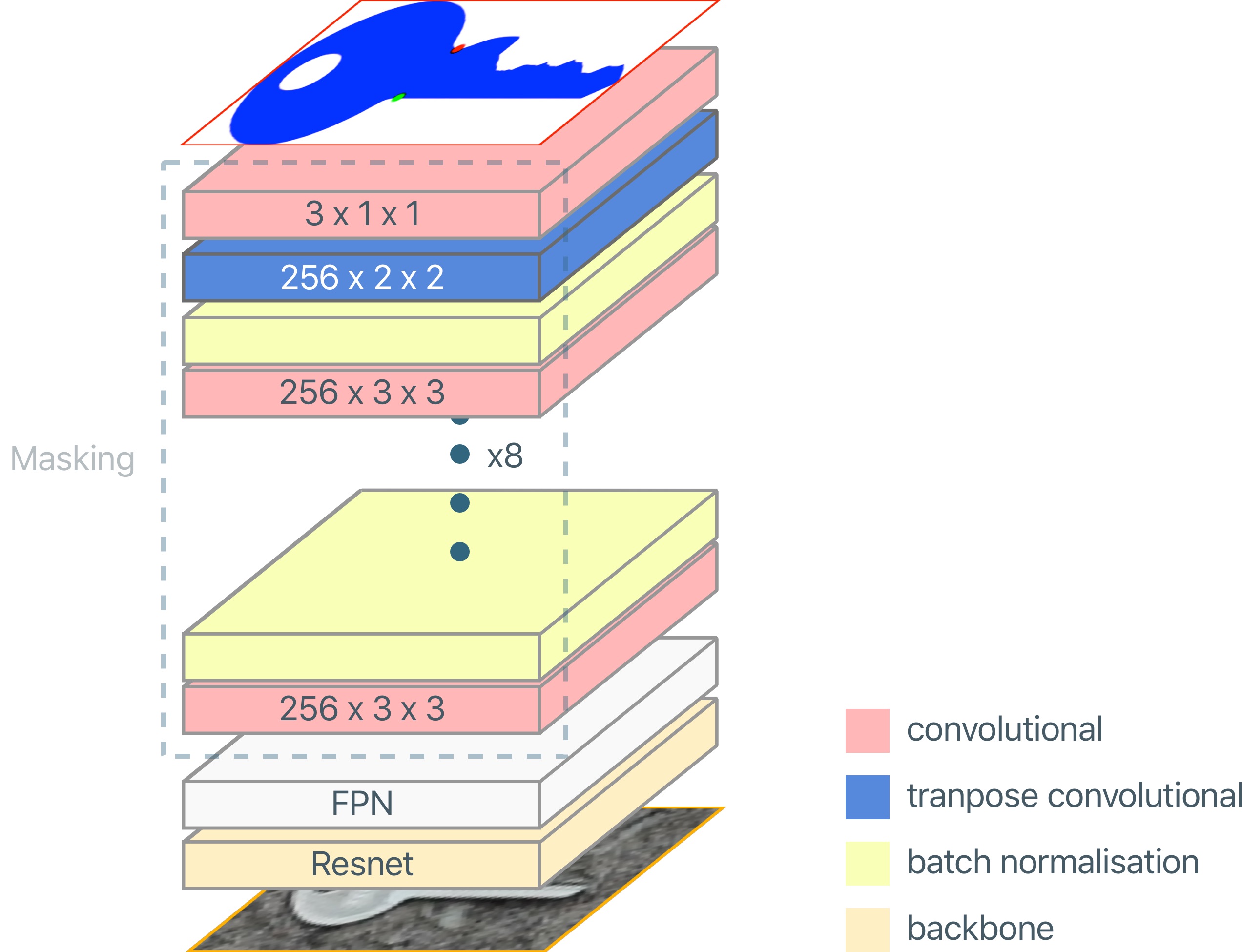}\textcolor[rgb]{1,1,1}{....} \includegraphics[width=82pt,height=111pt]{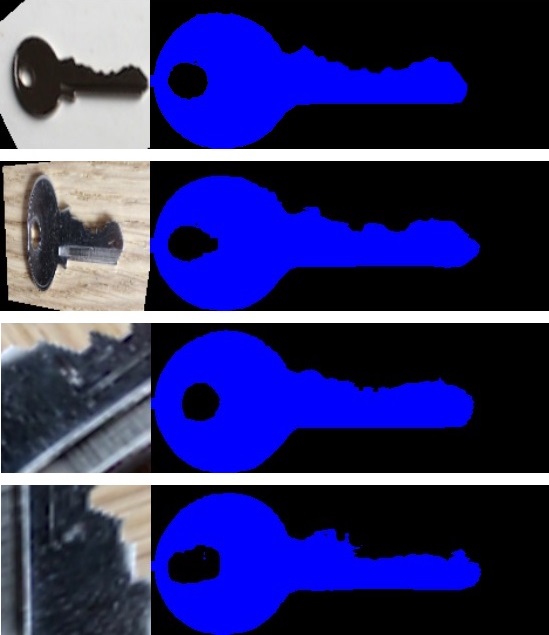}\textcolor[rgb]{1,1,1}{..} \includegraphics[width=127pt,height=71pt]{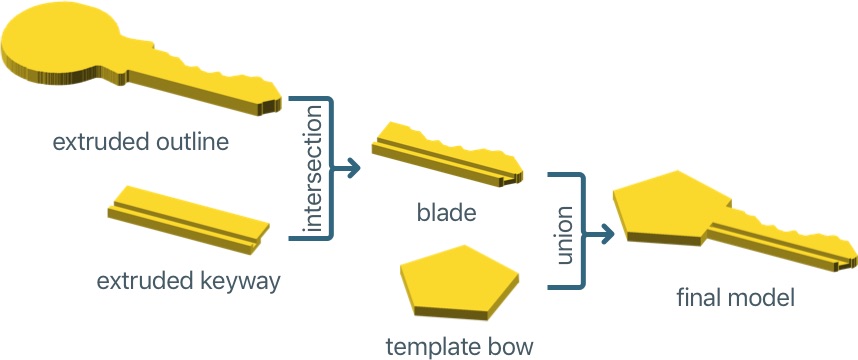}
\caption{{\textbf{Segmentation Architecture, Mask Examples and Key Modelling.} \textbf{\textit{(left)}} \textit{\textbf{Component Overview: }}Normalised key crops are fed through our modified Mask-RCNN head, where the additional deconvolutions and larger pool size provides us greater output resolution. \textbf{\textit{(middle)}} \textit{\textbf{Segmentation Examples after End-to-End Optimisation: }}Final pose normalisations and bitting segmentations after running test samples through the full end-to-end system. Masks are resized from $56\times56$ pixel output. Top 2 examples show valid bitting segmentation whereas others highlight cases where normalisation errors cause incorrect bitting segmentation. These incorrect segmentation look like plausable key masks but do not match the ground truth. \textbf{\textit{(right)}}~\textit{\textbf{Modelling Process: }}Separate components of the key that are either provided by the user or determined by the pipeline are fused together to create a printable STL file.}}
\label{fig:archseg}
\end{figure}

Figure \ref{fig:archseg} depicts how normalised patches are utilised to produce bitting masks inspired by Mask R-CNN \cite{He2017MaskRCNN}. However, in contrast to \cite{He2017MaskRCNN} and given  scale pre-normalisation, we only compute FPN layer $P2$ encoded at the highest resolution and use $8$ convolutions (rather than $4$) before de-convolution. We found this provides higher precision in mask outputs when paired with a pool size of $28$, resulting in a final mask size of $56\times 56$ pixels as seen in Figure~\ref{fig:archseg}. In order to accurately align the output mask with the keyway, we define ground truth masks with small activation areas of two key points as further pixel classes to train on -- following  \cite{He2017MaskRCNN}. We use DeepKey Dataset B at a LR of $0.001$ and SGD with momentum ($\lambda=0.9$) for training. LR was reduced by a factor $10$ after $400$ epochs before running further $400$ epochs. As given in \cite{He2017MaskRCNN}, our masking loss was average binary cross entropy loss with a per-pixel sigmoid on final layer logits.

    \subsection{3D Model Generation and Printing}
    \label{sec:implementation_and_methodology:3d_Scripting}
    
    The estimated key mask, key points, keyway profile and real-world keyway height are used for 3D key model generation. Binarisations of the bitting mask are transformed into a series of key boundary points. Using this description, this bitting and the known keyway can be extruded along orthogonal axes and key point locations are aligned (scaled+translated) with the keyway via scripts~\cite{openscad}, where the final key blade is  constructed by union and attachment of a standard key bow  before outputting an STL file printable using~\cite{ultimaker2plus}. 

    \subsection{Final Optimisation via an End-to-End Pipeline }
    \label{sec:implementation_and_methodology:end_to_end_implementation}

    After training all subcomponents as described, we then progress to the end-to-end training of components by forwarding data generated by each subcomponent as training data to subsequent components in order to optimise the overall system. This two-step process of bootstrapping each subcomponent's weights from separately provided ground truth mitigated the effect of early stage errors.
 
During end-to-end training, we only generate training data from the DeepKey Dataset B adjusting the detection component to take input sizes of $1024\times 1024$ (up from the $224\times 224$ standard of \cite{AlexKrizhevsky2012ImageNetCW}) due to resolution requirements of the later stages. Additionally, we use input batches of 2 and a detection proposal limit of 32 so that later stages experience an effective batch size of 64 because of GPU size limitations -- all other training parameters are inherited from the definitions provided in each subcomponent's section. Inputs are processed end-to-end for $200$ epochs to train with frozen backbones, followed by a further $800$ epochs at $1/10$ LR, and a final $800$ epochs with unfrozen backbones and a batch size of only $2$ at a LR of $10^{-7}$.
    
In order to fit the end-to-end pipeline onto Blue Crystal 4 \cite{UoBcomputerCenterBC4} Nvidia P100 GPU nodes used for all optimisations of this paper, we opted to share the learned weights between the three backbone instances across the architecture. Despite weight sharing, performance improvements can still be recorded under our forwarding end-to-end paradigm.

\section{Results}
\label{sec:experiments}
This section discusses and evaluates in detail metrics that quantify the performance of each subcomponent and the overall DeepKey system.
For detection evaluation, six different types of test data arrangements will be used, all derived from withheld testset portions of the DeepKey Datasets: 1) first, 875 withheld original low-resolution images from Dataset A; 2) 875 images derived from the former via augmentation (reminder of Figure \ref{fig:marker}); 3) 889 withheld original high-resolution images from Dataset B; 4) 889 images derived from the former via augmentation; 5) 889 withheld original high resolution images from Dataset~B with marker frame removal; and 6) 889 images derived from the former via augmentation. All subsequent component evaluations utilise~Dataset~variation~3. 

\begin{figure}[!ht]
\centering
\begin{minipage}{0.39\textwidth}
\centering
        \begin{tabular}{c|c|c}
        Test Dataset & Augmented & AP \\
        \hline
        (1) A & & 0.955 \\
        (2) A & \checkmark & 0.771 \\
        (3) B & & \textbf{0.997} \\
        (4) B & \checkmark & 0.978 \\
        (5) B noframe & & 0.977 \\
        (6) B noframe & \checkmark & 0.911 \\
        \end{tabular}
\end{minipage}\textcolor[rgb]{1,1,1}{-}
\begin{minipage}{0.59\textwidth}
        \includegraphics[width=211pt,height=135pt]{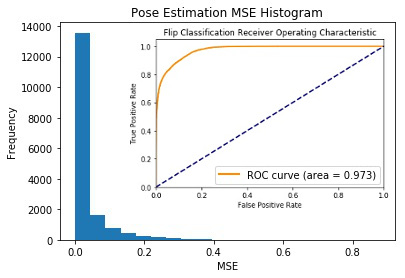}
\end{minipage}
\caption{\textbf{Detection and Normalisation Visualisations and Metrics.} \textbf{\textit{(left)}} \textit{\textbf{Detection AP Metrics: }}Results are shown for all six test cases covering original test data, geometric augmentations, as well as  'noframe' tests with augmentation of colours of all marker pixels and those outside the marker black. \textbf{\textit{(right)}} \textit{\textbf{Pose Normalisation Metrics: }}Pose normalisation results in low parameter value MSE achieved on the DeepKey TestSet B, where flip classification achieves an AUC of $0.973$. }
\label{fig:resdet2}
\end{figure}

\begin{figure}[!hb]
\centering
\begin{minipage}{0.48\textwidth}
        \includegraphics[width=171pt,height=110pt]{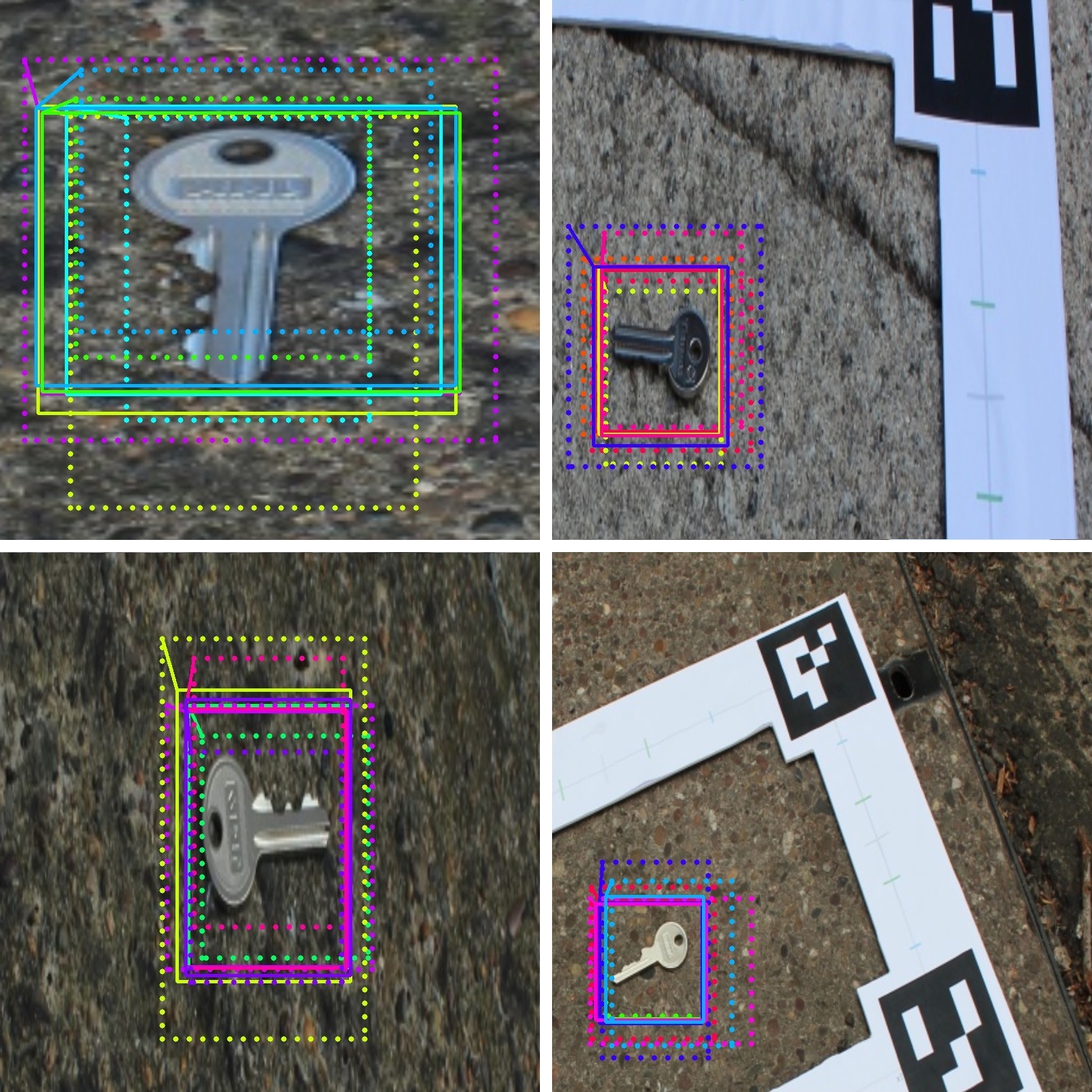}
\end{minipage}\textcolor[rgb]{1,1,1}{....}
\begin{minipage}{0.48\textwidth}
        \includegraphics[width=171pt,height=125pt]{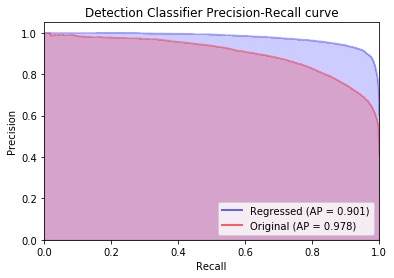}
\end{minipage}
\caption{\textbf{Detection Head Performance.} \textbf{\textit{(left)}} \textit{\textbf{Correct Proposal Samples:}} Four input images and their detection head bounding box regressions. Dotted boxes indicate original proposals from the RPN and are depicted with their regressed solid box counterparts by similarly coloured lines. In the featured samples, the firstly somewhat misaligned RPN proposals are regressed to common bounding boxes that correctly bound their target keys. \textbf{\textit{(right)}} \textit{\textbf{Precision-Recall Metric:}} Precision-recall curve for the first ($AP=0.901$) and then the regressed final  detection head proposals ($AP=0.978$ as reported in Figure \ref{fig:resdet2}) when run on augmented images derived from DeepKey Testset~B.}
\label{fig:resdet}
\end{figure}

Marker frame removal is applied to guarantee full independence from markerboard presence during tests. Whilst augmentations (see Section \ref{sec:main_body}) allow for validation using a larger number of geometric transforms than found in the data gathered, in order to ensure we are not applying augmentations to manipulate object detection results in our favour, we also compare to non-augmented data. Figure \ref{fig:resdet2} (left)  shows an overview of key detection performance results for all six test data arrangements. Figure \ref{fig:resdet} then exemplifies detection and quantifies localisation regression. It provides a precision-recall analysis of improvements provided by localisation regression.

\begin{figure}[t]
\centering
\includegraphics[width=324pt,height=125pt]{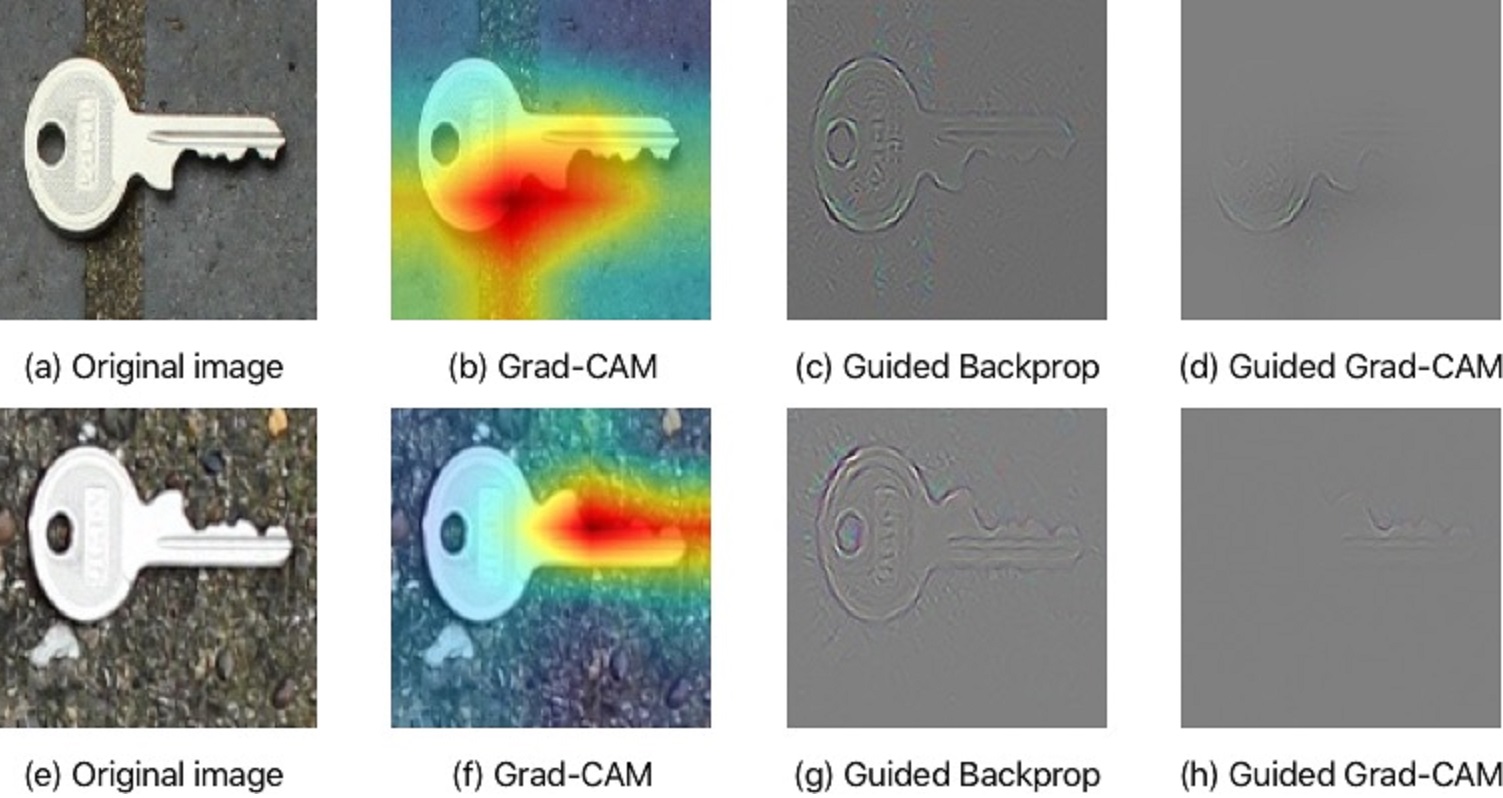}
\caption{\textbf{Flip Classification Visualisation.} \textbf{\textit{(a,e)}} \textit{\textbf{Original Input Patches:}} RGB normalised key crops of the flipped and unflipped classes. \textbf{\textit{(b,f)}} \textbf{\textit{Grad-CAM Heatmaps:}} Overlayed on top of original images showing areas of interest to the final convolutional layer before fully connected layers and predictions. Note how on the flipped key the shoulder of the key is seen as interesting to the final layer whereas on the unflipped key, the key bitting is used. \textbf{\textit{(c,g)}} \textit{\textbf{Guided Back-Propagation:}} Visualisation of features important to the  prediction. \textbf{\textit{(d,h)}} \textit{\textbf{Guided Grad-CAM:}} Combining Grad-CAM with guided back-prop to highlight potentially class-discriminative features. }
\label{fig:gradcam}
\end{figure}
It can be seen that key-class object detection produces the highest AP~of~$0.997$ upon the original DeepKey Testset B. The system achieves a lower AP when full augmentation is applied. Our results show only a negligible decrease in AP when removing marker information. Thus, the bias towards images containing markers as introduced in training is small -- key detection AP on the original DeepKey Testset B reduces by $2.00\%$ from $AP=0.997$ to $AP=0.977$ upon marker frame removal from test instances. 
Figure \ref{fig:resdet2} (right)  shows an overview of pose normalisation performance results, where the process can be seen depicted in~Figure~\ref{fig:archnorm}~(right) key samples are successfully normalised and flipped. We record the MSE across all values of the transformation matrices generated by pose normalisation inference to produce a histogram showing the error distribution. In addition, we produce a receiver operating characteristic (ROC) plot for flip classification.  Our flip classification network operates with an AUC of $0.973$. Via the application of Gradcam \cite{Selvaraju2016GradCAMWD}, it can be seen that, as depicted in Figure \ref{fig:gradcam}, the network activations of this component focus indeed on key bitting and shoulder during flip classification, as required to determine orientation. Transformation estimation performs as described in Figure \ref{fig:resdet2} (right), however, a small absolute error can of course still result in a clearly visible transformation error, as depicted by non-horizontal keys in Figures \ref{fig:archnorm} (right).

\begin{figure}[t]
\centering
\includegraphics[width=344pt,height=30pt]{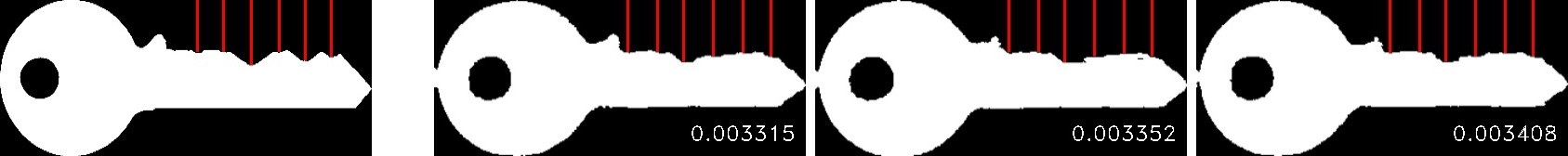} \includegraphics[width=344pt,height=30pt]{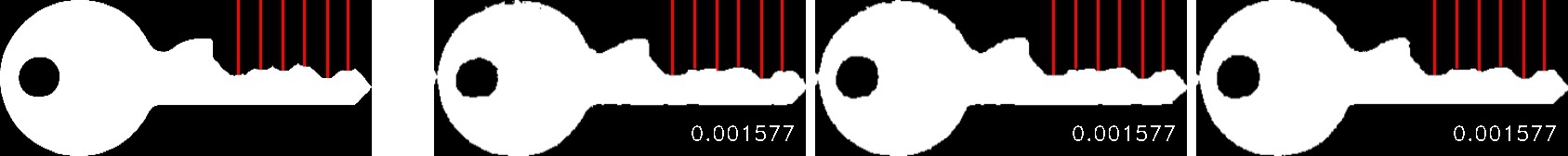}
\includegraphics[width=344pt,height=30pt]{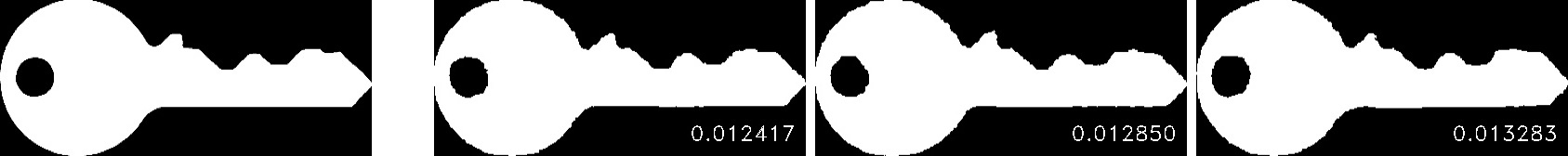} \includegraphics[width=344pt,height=30pt]{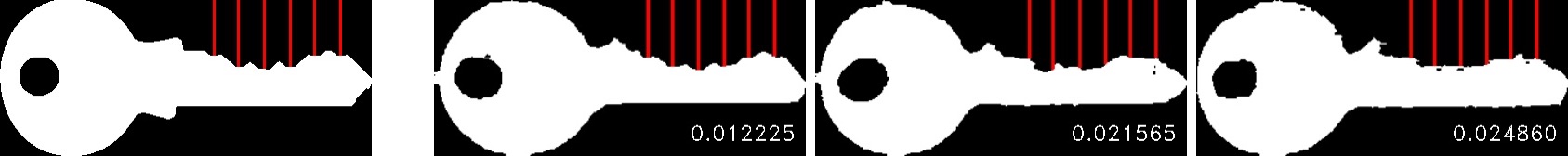}
\caption{\textbf{Bitting Segmentation Error Visualisation.} Ground truth masks in left column, segmentation examples of this key on the right. Per-mask errors are given on each mask. \textbf{\textit{(top to bottom)}} \textit{\textbf{MPE: }}Error is calculated as max difference between ground truth cut height and measured segmentation cut height at virtual pin locations. This metric favours keys with shallow bittings as segmentations tend to smooth. \textit{\textbf{Mean: }}Error is calculated as mean difference between ground truth cut height and measured cut height at virtual pin locations. We find this metric scores best on validation keys of a similar type to those in the training set, which exhibit little variation. \textit{\textbf{IoU: }}Error is the pixelwise intersection between mask and ground truth, divided by the area of the mask. Segmentations with low edge noise produce the lowest error scores with this metric. \textit{\textbf{Unseen MPE: }}Same as \textit{MPE}, but with an unseen key, not in dataset A or B. }
\label{fig:bitting_segmentation}
\end{figure}

When evaluating the bitting segmentation, we take an approach contrary to the standard in image segmentation: rather than only calculating the pixel mask IoU as in \cite{He2017MaskRCNN}, we also focus our evaluation on the bitting of the keys by casting rays at manually annotated locations required for pin lifting in a lock. These \textit{virtual pins} emulate the success criteria in the real-world where lock pins must be raised or lowered to a satisfactory height. Figure \ref{fig:bitting_segmentation} comprehensively visualises the metrics we use to evaluate the bitting segmentation, utilising three different metrics: (top row) the `max pin height' (MPE) error reflects the largest difference across all cuts on a key, and is the most practical metric as reality requires that not one pin produce error beyond some threshold; (second row) `mean pin height' error takes the mean across all pins per key, producing a conceptual metric of how far the bitting of a key is from ground truth; (third row) `pixel IoU' is included for the sake of comparison.

The practically most relevant metric is that of MPE: Figure \ref{fig:bitting_segmentation} (bottom) includes a mask never seen in training data (outside DeepKey Datasets A and B) with an MPE of $0.0122$ -- we use example output to print a physical sample and successfully open its target lock as shown in Figure \ref{fig:res}. 
For this test, we estimate physical operation capable of unlocking the target when used with a key showing a segmentation error of $MPE <= 0.012$, however, we note that higher quality locks will feature lower bitting error tolerances. Across  our validation set, mean MPE is $0.039$, far higher than the  quality example  given in Figure~\ref{fig:bitting_segmentation}~(top).

\section{Physical Proof-of-Concept}
\label{sec:poc}
\begin{figure}[!t]
\centering
\includegraphics[width=221pt,height=190pt]{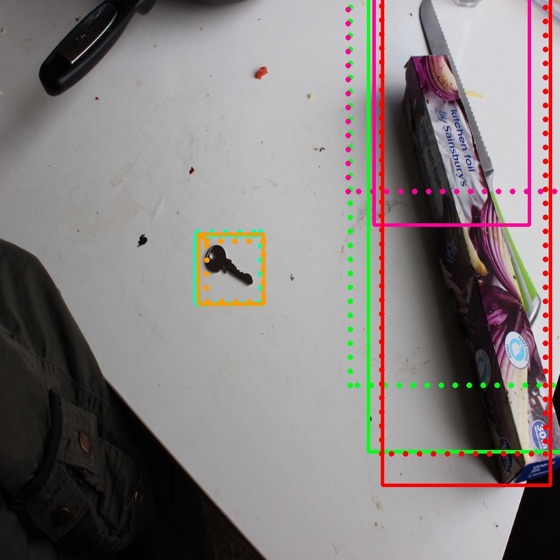}\textcolor[rgb]{1,1,1}{.}\includegraphics[width=100pt,height=190pt]{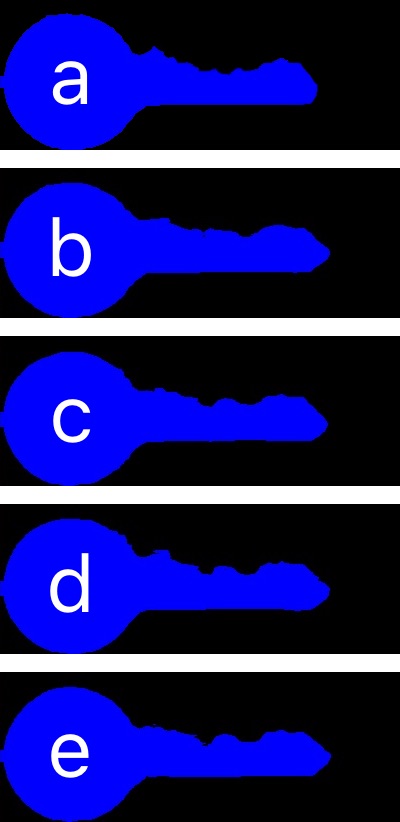}
\includegraphics[width=324pt,height=130pt]{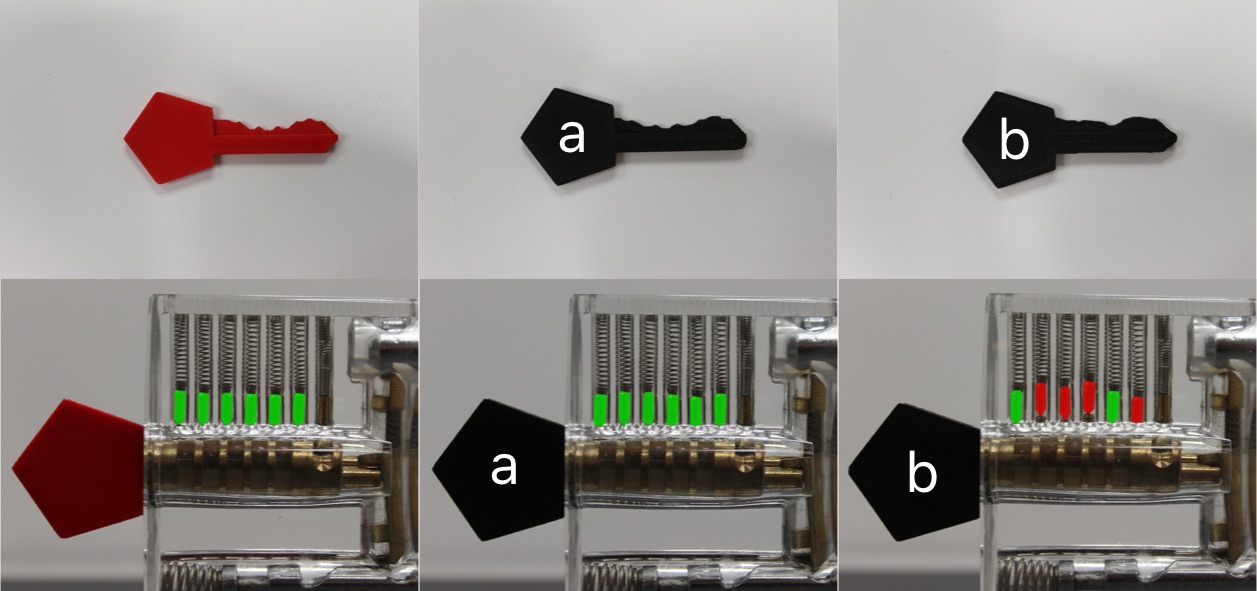}
\caption{\textbf{Physical Key Prediction Example.} \textbf{\textit{(top-left)}} \textit{\textbf{Example of Unseen Image: }}Sample image from the unseen DeepKey Dataset C, without the inclusion of any marker frames.  The selected key detection undergoes normalisation and is rectified, which in turn allows segmentation to successfully take place. \textbf{\textit{(top-right)}} \textit{\textbf{Sampled Segmentations: }}$5$ manually selected samples from the end-to-end segmentation output used to print physical keys for tests. \textbf{\textit{(bottom)}} \textit{\textbf{3D Printed Key Opens Lock: }}Red ground truth key raises all lock pins to their valid positions. Print (a) resulting from the above end-to-end system output also raises all lock pins to sufficient levels, although they are visibly slightly different from those of the ground truth key. A secondary print (b) exhibits too large deviance in bitting  to open the lock. }
\label{fig:res}
\end{figure}

Consequently, according to the result statistics, only a small proportion of physical replicas generated are expected to work  in opening the target lock. To provide a proof of concept that real-world locks can indeed be opened using the system, we evaluate the full pipeline empirically  using a tiny, new set of test images (withheld DeepKey Dataset C) of a hitherto, unseen key exemplified in the top left image of Figure \ref{fig:res}. We manually select the top $5$ end-to-end masks -- three of these masks can be seen in Figure \ref{fig:bitting_segmentation} (bottom). From this test set of 5, we find that only $1$ is capable of unlocking the target lock as shown in Figure \ref{fig:res}.~Although the heights of the pins differ from the ground truth key, the lock still operates correctly due to in-built tolerances as explained in the results section. 
We would expect truly high-quality locks to reject such a key.
For this proof-of-concept study many prints were needed to yield one that opens the target~lock. Nevertheless, the study shows that given an image of appropriate quality, the system \textit{is} after possibly many trials eventually capable of producing a valid, lock-opening key from a single visual image. Further analyses of the detailed conditions that lead to images and prints of sufficient quality to successful open locks with the system would be an important step, however, this is outside the scope of this paper and would require significant further work.

\section{Reflection, Potential Societal Impact, and Conclusion}
\label{sec:conclusion}

The  usability of basic key imagery for the production of  key models that are potentially capable of unlocking a physical target lock, be that via a traditional vision system~\cite{Laxton2008ReconsideringPK} or via a deep learning approach such as DeepKey, raises various questions about any potential impact on physical security and society overall.

 Whilst applications for legitimate owners to visually backup their keys could mitigate accidental loss, one also has to consider the case of assailants taking a series of still RGB pictures of a key through a window and printing replicas using a potentially mobile 3D printer. Our results show that DeepKey, as described, is of limited use in such a scenario since  multiple key models and prints are likely to be required, as not every model will be valid, and the number of attempts needed will rise as the lock quality increases due to lower error tolerances. Thus, it is probable that many prints, possibly $10$s or even $100$s may be needed to produce a working key using DeepKey depending on the scenario. 

Whilst we believe that increased final {prediction resolution} trained via larger GPUs may be beneficial re improving accuracy, the potential for systems like DeepKey to cause a general threat to public lock-users is also hamstrung by the large variety of lock types available today. We assessed only the application of automated key model prediction to one type of Yale pin tumbler keys. 

Most importantly, however, everyone should consider that basic countermeasures such as avoidance of visual exposure or bitting-covering key rings are simple and effective ways to deny unwanted key imaging.  New architectures and approaches to lock-based security such as those featuring multi-sided 3D bittings may also be effective in minimising the risk of unwanted photographic capture. Magnetic locks in particular provide security properties outside the visual domain since the key's secret is encoded in the orientation of embedded magnets. Due to the lack in visual variance of such keys, any visual system would be entirely ineffective in capturing the key's secret information. 

We hope that the publication of DeepKey can inspire research into countermeasures and legitimate security applications, whilst also acting as an early warning:  academia, lock producers, authorities and the general public must be alerted to the growing potential of  deep learning driven visual attacks  in an unprepared physical security~world.

        \bibliographystyle{splncs04}
        \bibliography{egbib}

\end{document}